\documentclass[runningheads]{llncs}

\usepackage{eccv}

\usepackage{eccvabbrv}
\usepackage[dvipsnames]{xcolor}
\usepackage{algorithm}
\usepackage{algpseudocode}
\usepackage{amsmath}
\usepackage{makecell}
\usepackage{multirow}
\usepackage{colortbl}
\usepackage{adjustbox}
\usepackage{utfsym}
\usepackage{graphicx}
\usepackage{booktabs}
\usepackage[accsupp]{axessibility}  

\usepackage{hyperref}

\usepackage{orcidlink}

\begin{document}

\title{Rethinking Video Deblurring with Wavelet-Aware 
Dynamic Transformer and Diffusion Model} 

\titlerunning{Video Deblurring with Transformer and Diffusion Model}

\author{Chen Rao\inst{1}\textsuperscript{\dag}\orcidlink{0009-0008-8059-3438} \and
Guangyuan Li\inst{1}\textsuperscript{\dag}
\orcidlink{0000-0002-2938-527X} \and
Zehua Lan\inst{1}\orcidlink{0000-0001-7486-2103} \and
Jiakai Sun\inst{1}\orcidlink{0000-0003-2894-9456} \and
Junsheng Luan\inst{1}\orcidlink{0009-0009-2902-8086} \and
Wei Xing\inst{1}\textsuperscript{*}\orcidlink{0000-0002-7994-6725} \and
Lei Zhao\inst{1}\textsuperscript{*}\orcidlink{0000-0003-4791-454X} \and
Huaizhong Lin\inst{1}\textsuperscript{*}\orcidlink{0000-0001-6313-5349} \and
Jianfeng Dong\inst{2} \and
Dalong Zhang\inst{1}
}

\authorrunning{C. Rao et al.}

\institute{College of Computer Science and Technology, Zhejiang University \and Zhejiang Gongshang University}

\renewcommand{\thefootnote}{}
\footnotetext{\textsuperscript{\dag} co-first authors, equal contribution; \textsuperscript{*} Corresponding author}

\maketitle

\begin{abstract}
Current video deblurring methods have limitations in recovering high-frequency information since the regression losses are conservative with high-frequency details. Since Diffusion Models (DMs) have strong capabilities in generating high-frequency details, we consider introducing DMs into the video deblurring task. However, we found that directly applying DMs to the video deblurring task has the following problems: (1) DMs require many iteration steps to generate videos from Gaussian noise, which consumes many computational resources. (2) DMs are easily misled by the blurry artifacts in the video, resulting in irrational content and distortion of the deblurred video. To address the above issues, we propose a novel video deblurring framework VD-Diff that integrates the diffusion model into the Wavelet-Aware Dynamic Transformer (WADT). Specifically, we perform the diffusion model in a highly compact latent space to generate prior features containing high-frequency information that conforms to the ground truth distribution. We design the WADT to preserve and recover the low-frequency information in the video while utilizing the high-frequency information generated by the diffusion model. 
Extensive experiments show that our proposed VD-Diff outperforms SOTA methods on GoPro, DVD, BSD, and Real-World Video datasets.
The codes will be available at \url{https://github.com/Chen-Rao/VD-Diff}.
\end{abstract}

\section{Introduction}
\label{sec:intro}

Portable devices are widely used to capture videos. However, camera shake and fast movement of objects lead to undesirable blur in video. To reduce the impact of video blur, researchers have done a lot of work on video deblurring. Traditional methods are mainly based on hand-crafted assumptions and priors, which are difficult to design. Besides, inappropriate priors will reduce the quality of video deblurring. 

In the past decade, video deblurring has witnessed significant progresses with the development of deep learning. Early deep learning methods mainly use CNN-based models \cite{hyun2015generalized, zhang2018adversarial, 2019EDVR, 2020CDVDTSP, hyun2017online, nah2019recurrent, 2020ESTRNN} to extract video features for capturing the spatio-temporal information between frames. Nevertheless, these approaches have limitations in modelling long-range spatial dependencies and capturing non-local self-similarities.

Recently, Transformer has been employed in video deblurring tasks \cite{wang2022uformer, NAFNet, FGST, 2022rvrt} to address limitations in CNN-based methods, which can establish long-range spatial dependencies and capture non-local spatial information.
Although Transformer-based methods exhibit impressive performance, they still have the following issues.
On the one hand, the methods using standard global Transformer \cite{dosovitskiy2020image} tend to deblur videos based on low-frequency details and have limited perception of high-frequency information. 
On the other hand, when the local window-based Transformer \cite{wang2022uformer} is used to reduce the computational cost, self-attention is calculated within a location-specific window, resulting in a limited receptive field. Consequently, it may ignore some key elements of similar and clearer scene patches in the spatiotemporal neighborhood. 
Therefore, the local window-based self-attention fails to capture valuable high-frequency information from the spatiotemporal neighborhood, resulting in the loss of high-frequency information.

Very recently, diffusion models (DMs) have shown their outstanding capabilities in video synthesis tasks \cite{rombach2022high, esser2023structure, blattmann2023align, ruan2023mm, Liu_2023_WACV, Yu_2023_CVPR, luo2023videofusion}, which can denoise pure Gaussian noise into high-quality video with abundant high-frequency details. 
Therefore, we consider employing DMs in video deblurring to restore high-frequency information.
However, directly applying DMs to video deblurring tasks has the following problems.
First, DMs demand a large number of iteration steps to generate videos from Gaussian noise, which costs a lot of computational resources.
Second, DMs are sensitive to motion blur in videos, resulting in irrational content and distortion of the deblurred video.

To solve the above problems, we propose a novel method called VD-Diff, which combines our proposed Wavelet-Aware Dynamic Transformer (WADT) and diffusion model for video deblurring. 
Specifically, 
we employ the diffusion model in a highly compact latent space to generate prior features containing high-frequency information that conforms to the ground truth distribution, which facilitates video deblurring. Applying DM in a highly compressed latent space can significantly reduce the number of denoising iteration steps.

In addition, we design WADT to retain and restore the low-frequency information in the video while exacting the high-frequency information from the prior features generated by the diffusion model, which ensures the deblurred video is undistorted.
WADT first decomposes the blurred video frames into the approximation coefficient and detail coefficients by Wavelet Transform (WT) and extracts low-frequency information from the approximation coefficient by Wavelet-Aware Dynamic Transformer Layers (WADTL). 
Meanwhile, WADT employs WADTL to fuse the prior features generated by DM containing high-frequency information to generate artifact-free and distortion-free videos.
Furthermore, we introduce a Wavelet-based Bidirectional Propagation Fuse (WBPF)  module in WADT to fully use the spatio-temporal information between frames, which ensure that the video is undistorted in the spatio-temporal dimension.
Our main contributions are as follows:

(1) We propose a novel method combining the Wavelet-Aware Dynamic Transformer (WADT) and the diffusion model for video deblurring. To the best of our knowledge, our study is the first to employ DM for video deblurring, which requires only a few iteration steps to restore artifact-free and distortion-free videos.

(2) Our diffusion model can generate prior features containing high-frequency information that conforms to the GT distribution, which facilitates video deblurring.

(3) Our proposed WADT can capture low-frequency information while utilizing the prior features generated by the DM to supplement high-frequency information, which further enhance the quality of deblurred video. 

(4) Extensive experiments show that our proposed VD-Diff outperforms SOTA methods on GoPro, DVD, BSD and Real-World Video datasets.

\begin{figure*}[t!]
  \centering
   \includegraphics[width=\linewidth]{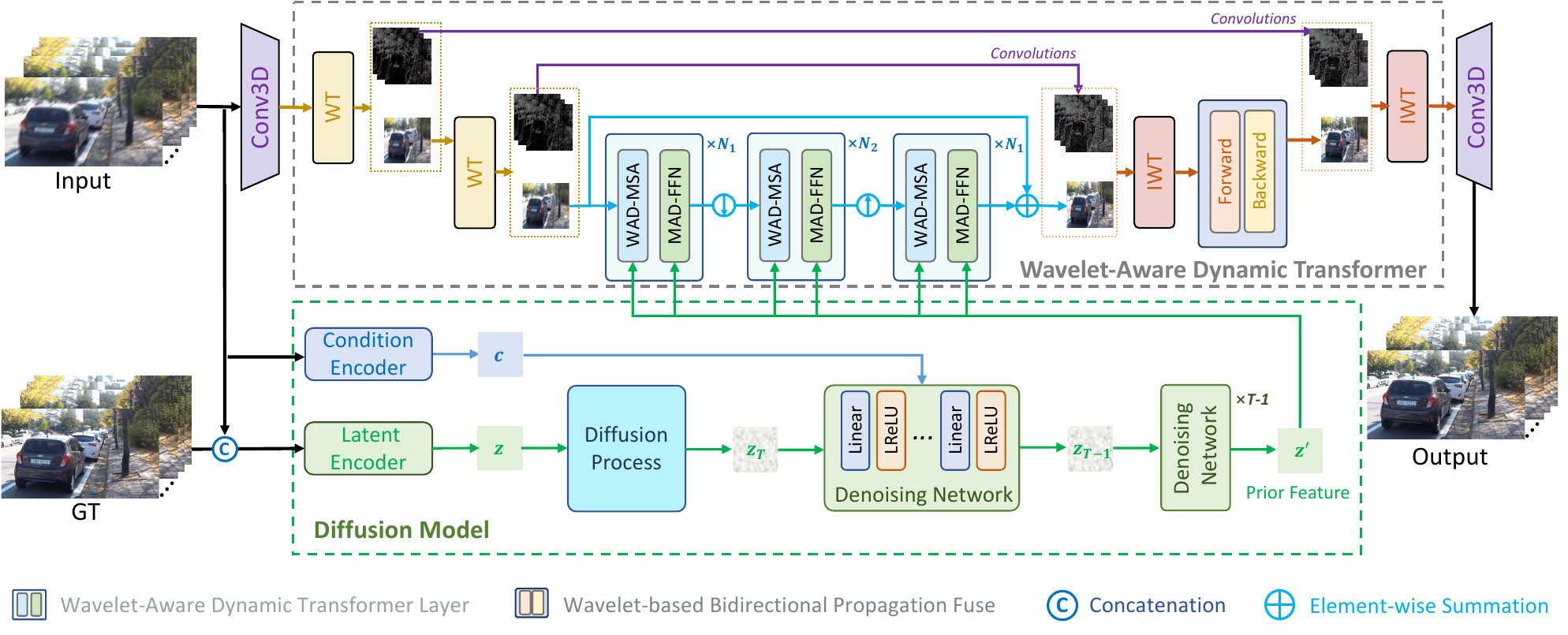}
   \caption{The overall architecture of the proposed VD-Diff, which consists of Wavelet-Aware Dynamic Transformer (WADT) and Diffusion Model (DM). Specifically, WADT adopts Wavelet Transform (WT) for feature separation, WADT Layer (WADTL) for deep feature extraction, and Wavelet-based Bidirectional Propagation Fuse (WBPF) for spatio-temporal information propagation between frames. The DM generates prior features to supplement high-frequency information for the deblurring process in WADTL.WT: Wavelet Transform. IWT: Inverse Wavelet Transform.}
   \label{main_struct}
\end{figure*}

\section{Related Work}
\subsection{Video Deblurring}
With the development of deep learning \cite{li2013efficient}, CNN is widely used in video deblurring. For instance, Zhong \emph{et al.} \cite{2020ESTRNN} propose residual dense blocks and a global spatio-temporal attention module for video deblurring; Pan \emph{et al.} \cite{2020CDVDTSP} develope a deep CNN model to estimate optical flow from intermediate latent frames. CNN-based methods have driven the development of video deblurring, but CNN has limitations in capturing long-range dependencies. To address these issues, some researchers applied the Transformer to video deblurring and achieved impressive results. For example, Lin \emph{et al.} \cite{FGST} propose a method to calculate self-attention by capturing key elements of adjacent frames through optical flow. Liang \emph{et al.} \cite{2022rvrt} propose a recurrent video restoration transformer to process local neighboring frames in parallel within a globally recurrent framework. However, the above Transformer-based methods have limited ability to capture high-frequency information, leading to the deblurred videos lacking details. 
\subsection{Diffusion Models}
DMs have shown excellent performance in the field of synthesis. LDM \cite{rombach2022high} reaches a near-optimal point between complexity reduction and detail preservation by applying DM in the latent space. Blattmann \emph{et al.} \cite{blattmann2023align} turns the image generator into a video generator by introducing a temporal dimension to the latent space diffusion model. Artbank \cite{zhang2024artbank} utilize textual embedding to learn the visual priors from the images, synthesizing highly realistic artistic stylized images. DMs have also shown strong capabilities in the field of image restoration. For instance, DiffIR \cite{xia2023diffir}, Hi-diff \cite{chen2023hierarchical} and DiffMSR\cite{li2024rethinking} use DM to generate prior representations for image restoration. Although some methods, such as DDIM \cite{ddim} and LDM \cite{rombach2022high}, have reduced the computational complexity of DM by changing the denoising process or compressing the image, using existing DM for video deblurring still requires unaffordable computational resources. To cope with the above issue, we apply DM in a highly compact latent space to reduce the iteration steps. We were the first to apply diffusion models to video deblurring, and our models achieved outstanding performance.

\subsection{Wavelet-based Restoration Methods}
Wavelet is widely used in low-visual restoration tasks \cite{zuo2023generative, chu2023rethinking, zhao2020uctgan, li2023migt, li2015non} as it can separate low-frequency and high-frequency information. For example, 
Huang \emph{et al.} \cite{huang2017wavelet} propose a wavelet-based CNN for multi-scale face super resolution.
WavTrans \cite{li2022wavtrans} combines wavelet transform with Transformer for feature extraction and fusion; DSTNet \cite{DSTNet} develops a wavelet-based feature propagation method to effectively propagate main structures from long-range frames for better video deblurring. 
Inspired by the above methods, we proposed Wavelet-Aware Dynamic Transformer, which can extract the global structure of the video and utilize the prior features generated by DM to restore high-frequency details.

\section{Methodology}
\subsection{Overall Architecture}
The overall framework of our propsed VD-Diff is shown in Fig. \ref{main_struct}, which consists of two parts: 
(1) Wavelet-Aware Dynamic Transformer (WADT); (2) Diffusion Model. Specifically, we first employ Diffusion Model to obtain a highly compact latent prior feature $z\in \mathbb{R}^{T\times C'}$, which contains supplementary high-frequency information. Note that $C'$ is the number of channels of latent features. Then, we employ WADT to deblur the blurry video frames $V_{blur}\in R^{T\times 3\times H\times W}$ into $V_{HQ}\in R^{T\times 3\times H\times W}$ with the help of the prior feature $z$.
\subsection{Wavelet-Aware Dynamic Transformer}
\begin{figure*}[t!]
  \centering
   \includegraphics[width=\linewidth]{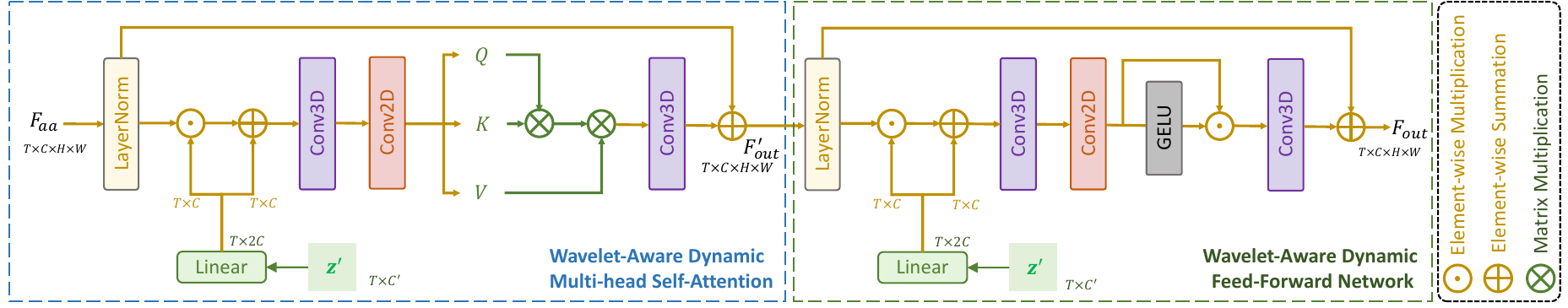}
   \caption{The illustration of Wavelet-Aware Dynamic Transformer Layer (WADTL), which consists of Wavelet-Aware Dynamic Multi-head Self-Attention (WAD-MSA) and Wavelet-Aware Dynamic Feed-Forward Network (WAD-FFN).}
   \label{WADTL}
\end{figure*}

As shown in the WADT part of Fig. \ref{main_struct}, we first utilize Conv3D to perform shallow feature extraction on the blurry video frames $V_{blur}\in R^{T\times 3\times H\times W}$ to obtain a global feature $F_{in} \in \mathbb{R}^{T\times C\times H\times W}$. 
Then, we use Wavelet Transform (WT) to decompose it into approximate coefficients $F_{a} \in \mathbb{R}^{T\times C/4\times H\times W}$ and detail coefficients $F_{d} \in \mathbb{R}^{T\times 3C/4\times H\times W}$ and again use WT on $F_{a}$ to obtain $F_{aa} \in \mathbb{R}^{T\times \hat{C}\times H\times W}$ and $F_{ad} \in \mathbb{R}^{T\times 3\hat{C}\times H\times W}$, where $\hat{C}=C/16$. Next, we employ WADTL to obtain the deblurred features $F_{out}$ from the approximate coefficients $F_{aa}$ and the prior feature $z'$. Then, we use WBPF to explore spatio-temporal information further. Finally, we use two IWTs to fuse features to obtain artifact-free and distortion-free videos $V_{HQ}$. We will elaborate on the structure of WADTL and WBPF as follows.\\
{\bf Wavelet-Aware Dynamic Transformer Layer.} As shown in Fig. \ref{main_struct}, we employ a stack of multiple WADT layer (WADTL) modules hierarchically to fuse approximate coefficients $F_{aa}$ and prior features $z'$. 
As shown in Fig. \ref{WADTL}, WADTL consists of a Wavelet-Aware Dynamic Multi-head Self-Attention (WAD-MSA) and a Wavelet-Aware Dynamic Feed-Forward Network (WAD-FFN). 

In the WAD-MSA part, given a approximate coefficients $F_{aa}$, we use the highly compact latent prior feature $z'$ to guide video restoration:
\begin{equation}
  \hat{F} = W_l^1z'\odot LN(F_{aa})+W_l^2z',
  \label{selfattention}
\end{equation}
where $\odot$ indicates element-wise multiplication, $LN$ denotes layer normalization \cite{ba2016layer}, $W_l$ represents linear layer. Then, we utilize a 3D convolution layer and a 2D convolution layer to project $\hat{F}$ into query $Q=W_2^QW_3^Q\hat{F}$, key $K=W_2^KW_3^K\hat{F}$, value $V=W_2^VW_3^V\hat{F}$ where $W_3$ is the channel-wise 3D convolution and $W_2$ is the depth-wise 2D convolution. Note that we use 3D convolution in the $T$, $H$, and $W$ dimensions and the same 2D convolution for the $T$ features in the temporal dimension $T$. 
Next, we perform the attention calculation:
\begin{equation}
  \mathcal{F}_{out}' = W_a\hat{V}\cdot Softmax(\hat{K}\cdot \hat{Q}/\gamma)+F_{aa},
  \label{WA-MSA}
\end{equation}
where $\gamma$ is a learnable scaling parameter.

\begin{figure}[t!]
  \centering
   \includegraphics[width=\linewidth]{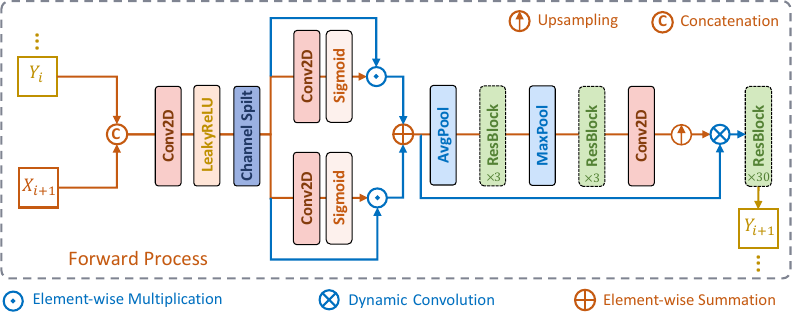}
   \caption{The structure of the Forward Process in the WBPF. The Backward Process has the same network structure as the Forward Process, but the direction of information propagation between frames is opposite.}
   \label{WBPF}
\end{figure}

In the WDA-FFN part, we first integrate the prior $z'$ into $\mathcal{F}_{out}'$ to obtain $\hat{\mathcal{F}}_{out}'$. And then we use Conv3D to aggregate temporal information and use Conv2D to aggregate spatial information from neighboring pixels.
Besides, we employ a gating mechanism to obtain more valuable information. 
The overall process of WDA-FFN can be described as:
\begin{equation}
  F_{out} = G(W_2^1W_3^1\hat{\mathcal{F}}_{out}^{\prime})\odot W_2^2W_3^2\hat{\mathcal{F}}_{out}^{\prime}+\hat{\mathcal{F}}_{out}^{\prime},
  \label{WA-FFN}
\end{equation}
where G denotes Gaussian Error Linear Units (GELU).\\
\noindent {\bf Wavelet-based Bidirectional Propagation Fuse.} Given the generated features $\left\{X_i\right\}_{i=1}^N$ by the first Inverse Wavelet Transform (IWT), to explore useful contents from the features of other frames, exiting methods usually simply stack $\left\{X_i\right\}_{i=1}^N$ or the alignment results of $\left\{X_i\right\}_{i=1}^N$. However, directly stacking frames without accurately estimating the features of $\left\{X_i\right\}_{i=1}^N$ or the alignment results of $\left\{X_i\right\}_{i=1}^N$ can only partially exploit the valuable information between frames and may also cause the propagation of erroneous information between frames. To further capture more useful long-range spatio-temporal information and reduce the influence of inaccurate information, we propose Wavelet-based Bidirectional Propagation Fuse (WBPF). 
As shown in Fig. \ref{main_struct}, WBPF consists of a forward process and a backward process.

We take the forward process as an example to introduce the WBPF module in detail, as shown in Fig. \ref{WBPF}.
First, we connect $Y_i$ and $X_{i+1}$ to get $\Bar{X}$, where $Y_i$ is the output of the previous forward process and $X_{i+1}$ is the input of the current forward process.
We then use a convolution layer with LeakyReLU to project $\Bar{X}$ into $\hat{X}$. Then we split $\hat{X}$ into $\hat{X}_1,\hat{X}_2$ along the channel dimension. 
Next, we obtain the fused feature from $\hat{X}_1,\hat{X}_2$ by:
\begin{equation}
  F_{pro}=\hat{X}_1\odot S(W_1\hat{X}_1)+\hat{X}_2\odot S(W_2\hat{X}_2),
  \label{DTFF1}
\end{equation}
where $S$ denotes the Sigmoid function and $W$ is the $3\times3$ 2D convolution. We utilize $AvgPool$, $ResBlock$, $MaxPool$, and $Conv2D$ to better explore spatial information. The overall process can be described as follows:
\begin{equation}
  \Bar{Y}_{i+1}=F_{pro}\otimes upsample(P(F_{pro})),
  \label{DTFF2}
\end{equation}
where P represents the stack of $AvgPool$, $ResBlock$, $MaxPool$, $ResBlock$ and $Conv2D$. In the last, we employ 30 ResBlocks on $\Bar{Y}_{i+1}$ to obtain ${Y}_{i+1}$.

\subsection{Diffusion Model}
Based on conditional denoising diffusion probabilistic models\cite{ho2020denoising, tashiro2021csdi, rombach2022high, ozdenizci2023restoring}, the DM involves a forward diffusion process and a reverse denoising process. We first adopt the Latent Encoder (LE) on ground truth video and the Condition Encoder (CE) on blurry video to generate the ground truth feature $z\in R^{T\times 4C'}$ and the condition feature $c\in R^{T\times 4C'}$. Then, we use the diffusion model to generate the prior feature $z'\in R^{T\times 4C'}$ using $c\in R^{T\times 4C'}$ as the condition and $z\in R^{T\times 4C'}$ as the target. We detail the diffusion process as follows.

In the forward diffusion process, we transform $z$ into Gaussian noise by $T$ iterations, which can be described as follows:
\begin{equation}
  q(z_{t}|z_{t-1})=\mathcal{N}(z_t;\sqrt{1-\beta_t}z_{t-1},\beta_tI),
  \label{ForwardIteration}
\end{equation}
where $t=1,2,...,T$; $z_0=z$; $\beta_t\in(0,1)$ are hyperparameters that control the variance of the noise. Using the notation $\alpha_t=1-\beta_t,\Bar{\alpha}_t=\prod_{i=1}^t\alpha_i$ and through iterative derivation with reparameterization, Eq. (\ref{ForwardIteration}) can be written as:
\begin{equation}
    q(z_T|z_0)=\mathcal{N}(z_T;\sqrt{\Bar{\alpha}_T}z_0,(1-\Bar{\alpha}_T)I),
  \label{ForwardDiffusion1}
\end{equation}
that is
\begin{equation}
    z_T=\sqrt{\Bar{\alpha}_T}z_0+\sqrt{1-\Bar{\alpha}_T}\epsilon.
  \label{ForwardDiffusion2}
\end{equation}

In the reverse denoising process, DM sample a Gaussian random noise $z_T$ and then gradually denoise $z_T$ to $z_0$:
\begin{equation}
  q(z_{t-1}|z_t,z_0)=\mathcal{N}(z_{t-1};\mu_t(z_t,z_0),\Tilde{\beta}_tI),
  \label{Reverse}
\end{equation}
\begin{equation}
    \mu_t(z_t,z_0)=\frac{\sqrt{\Bar{\alpha}_{t-1}}\beta_t}{1-\Bar{\alpha}_t}z_0+\frac{\sqrt{\alpha_t}(1-\Bar{\alpha}_{t-1})}{1-\Bar{\alpha}_t}z_t.
  \label{ReverseNotation1}
\end{equation}
We can use Eq. (\ref{ForwardDiffusion2}) to represent $z_0$ in Eq. (\ref{ReverseNotation1}) and obtain:
\begin{equation}
    \mu_t(z_t,z_0)=\frac{1}{\sqrt{\alpha_t}}(z_t-\epsilon\frac{1-\alpha_t}{\sqrt{1-\Bar{\alpha}_t}}).
  \label{MuFinal}
\end{equation}
We remove the variance estimation in Eq. (\ref{Reverse}):
\begin{equation}
    z_{t-1}=\frac{1}{\sqrt{\alpha_t}}(z_t-\epsilon\frac{1-\alpha_t}{\sqrt{1-\Bar{\alpha}_t}}).
  \label{zfinal}
\end{equation}
Then the noise $\epsilon$ is the only uncertain variable. Following previous work \cite{ho2020denoising,rombach2022high,xia2023diffir}, we adopt a neural network to predict the noise $\epsilon$ on $z_t$, $c$ and $t$.

\subsection{Training and Inference Strategies}\label{sec:TrainingStrategy}
Inspired by \cite{xia2023diffir, chen2023hierarchical}, we divide the training process into three stages: the first is to train the Latent Encoder with Wavelet-Aware Dynamic Transformer for ground truth feature extraction, the second is to train the Diffusion Model to generate the prior feature, and the third is to jointly train DM and WADT to generate artifact-free and high-realistic videos.

\noindent \textbf{Training Stage One.} 
The first stage aims to train the Latent Encoder (LE) to extract valuable ground truth prior feature $z$ from ground truth (GT). 
We train LE and WADT as a whole model by inputting $z$ into WADTL instead of $z'$ and ignoring the condition encoder and the diffusion model in Fig. \ref{main_struct}. We employ L1 loss and Multi-scale Frequency Reconstruction (MSFR) \cite{fan2021multi} loss. L1 loss is defined as:
\begin{equation}
  L_1=\sum_{i=1}^N \lVert V_{HQ}^i-V_{GT}^i\rVert_1, 
  \label{L1Loss}
\end{equation}
where $V_{HQ}^i$ is the $i$-th frame of the deblurred HQ video and $V_{GT}^i$ is the $i$-th frame of the GT video.
%
The MSFR loss measures the L1 distance between multi-scale GT and deblurred frame in the frequency domain as follows:
\begin{equation}
  L_{MSFR}=\sum_{i=1}^N\sum_{k=1}^K\frac{1}{t_k}\lVert \mathcal{FT}(\hat{S}_k^i)-\mathcal{FT}(S_k^i)\rVert_1,
  \label{MSFRLoss}
\end{equation}
where $\mathcal{FT}$ denotes the Fast Fourier Transform that transfers image signal to the frequency domain, $K$ denotes the number of different scales and the $\hat{S}_k^i$ indicates the image signal of the $i$-th frame at the $k$-th scale. 
The final optimization function of stage one is determined as follows:
\begin{equation}
  L_{deblur}=L_1+\lambda L_{MSFR},
  \label{DeblurLoss}
\end{equation}
where we experimentally set $\lambda=0.1$.

\noindent \textbf{Training Stage Two.} 
In the second stage, we first use a Condition Encoder (CE), whose struct is the same as the Latent Encoder (LE), to generate condition feature $c$ from the blur video. Then, we train a diffusion model (DM) to generate prior feature $z'$ using condition feature $c$.

We run the complete $T$ iteration reverse processes Eq. (\ref{zfinal}) to generate prior feature $\hat{z}$. We denoted this loss as:
\begin{equation}
    L_{diff}=\lVert z'-\hat{z}\rVert_1.
\label{DMLoss}
\end{equation}

\noindent \textbf{Training Stage Three.} 
In Training Stage Two, we train the DM to generate $z'$ to approximate $z$. However, there inevitably is a slight deviation between the prior features $z'$ and the ground truth prior feature $z$. To make the DM and WADT work better together to acquire information from different frequency domains for video deblurring and improve the robustness of WADT to the prior feature $z'$, we jointly train DM and WADT in Training Stage Three. The training loss $L_{total}$ is given by the sum of the deblurring loss $L_{deblur}$ and the diffusion loss $L_{diff}$:
\begin{equation}
    L_{total}=L_{deblur}+L_{diff}.
\label{TotalLoss}
\end{equation}

\noindent \textbf{Inference.}
Given a blurry input video $V_{blur}\in R^{T\times 3\times H\times W}$, we first compress $V_{blur}$ into a condition feature $c\in R^{T\times 4C'}$. Then, the diffusion model generates the prior feature $z'$ from pure Gaussian noise $\epsilon$ and condition feature c. Finally, VD-diff deblurs the blurry video with $z'$ using WADT.

\begin{figure*}[t]
\scriptsize
\centering
\begin{tabular}{ccc}
\hspace{-0.15cm}
\begin{adjustbox}{valign=t}
\begin{tabular}{c}
\includegraphics[width=0.45\textwidth]{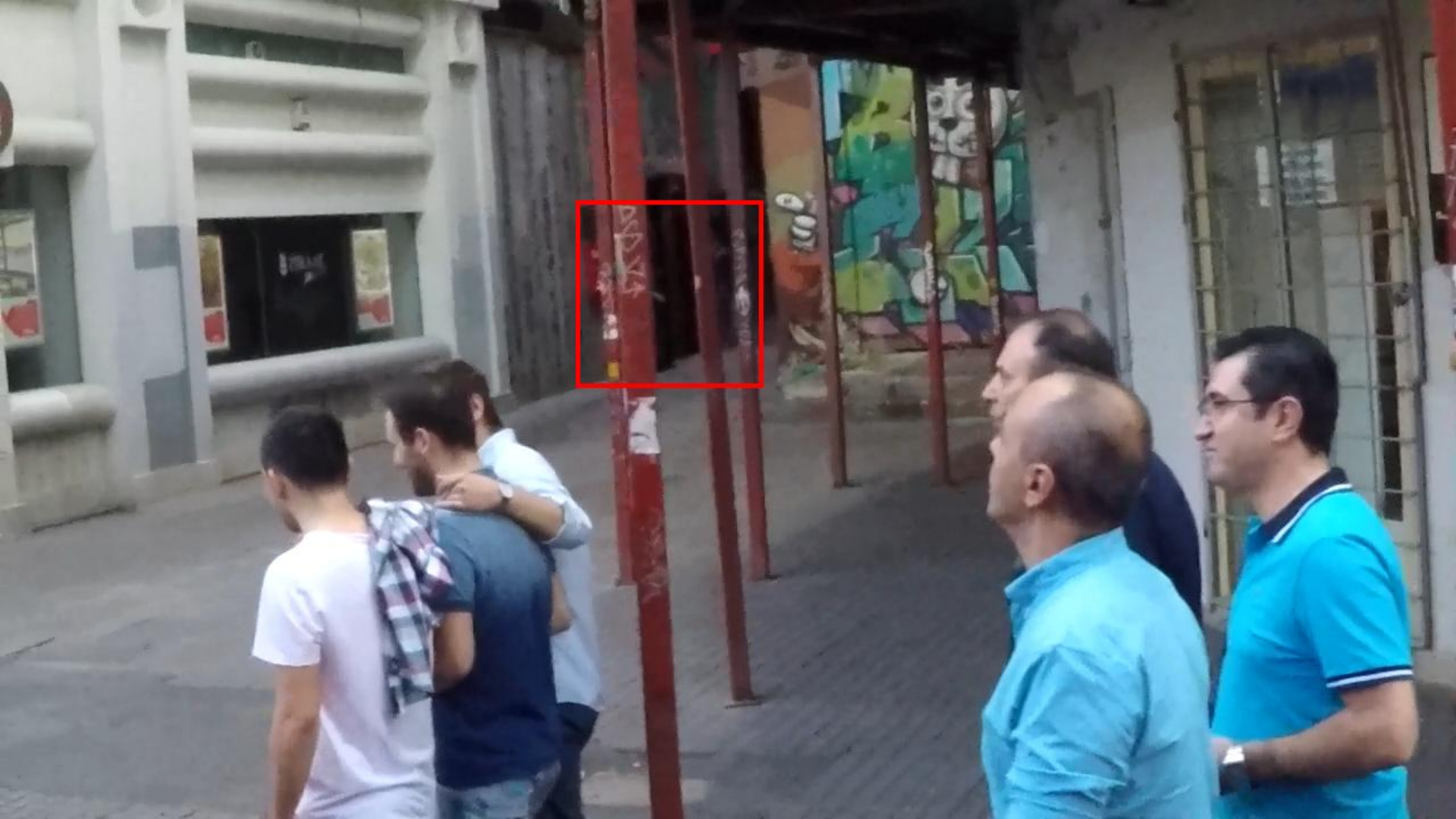}
\\
\end{tabular}
\end{adjustbox}
\hspace{-0.15cm}
\begin{adjustbox}{valign=t}
\begin{tabular}{cccccc}
\includegraphics[width=0.1\textwidth]{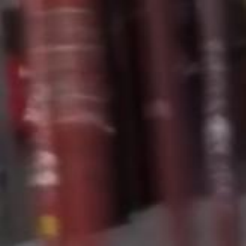} \hspace{-1mm}  &
\includegraphics[width=0.1\textwidth]{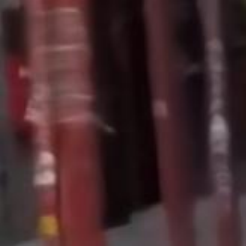} \hspace{-1mm}  &
\includegraphics[width=0.1\textwidth]{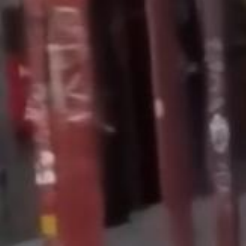} \hspace{-1mm} &
\includegraphics[width=0.1\textwidth]{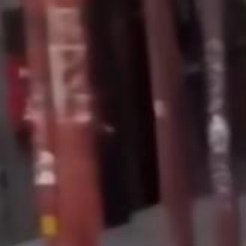} \hspace{-1mm} &
\includegraphics[width=0.1\textwidth]{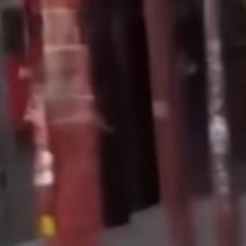} \hspace{-1mm}
\\
Blur \hspace{-1mm} &
EDVR \hspace{-1mm} &
MPRNet \hspace{-1mm} &
TSP \hspace{-1mm} &
SFE \hspace{-1mm}
\\
\includegraphics[width=0.1\textwidth]{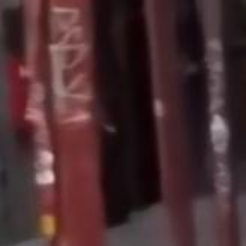} \hspace{-1mm} &
\includegraphics[width=0.1\textwidth]{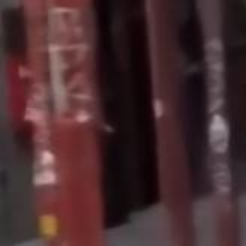} \hspace{-1mm} &
\includegraphics[width=0.1\textwidth]{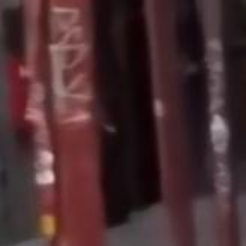} \hspace{-1mm} &
\includegraphics[width=0.1\textwidth]{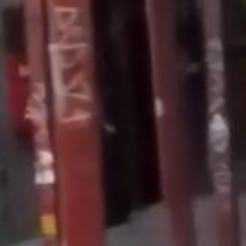} \hspace{-1mm} &
\includegraphics[width=0.1\textwidth]{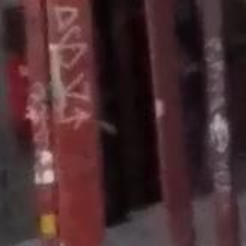}  \hspace{-1mm} 
\\ 
DSTNet \hspace{-1mm} &
FGST \hspace{-1mm} &
DSTNet-L \hspace{-1mm} &
Ours  \hspace{-1mm} &
GT \hspace{-1mm}
\\
\end{tabular}
\end{adjustbox}
\end{tabular}
\caption{Visual comparison on GoPro \cite{GoPro} dataset. The deblurred results of previous work still contain significant blur effects. Our method generates much clearer frames.}
\label{fig_gopro}
\end{figure*}

\begin{figure*}[t]
\scriptsize
\centering
\begin{tabular}{ccc}
\hspace{-0.15cm}
\begin{adjustbox}{valign=t}
\begin{tabular}{c}
\includegraphics[width=0.45\textwidth]{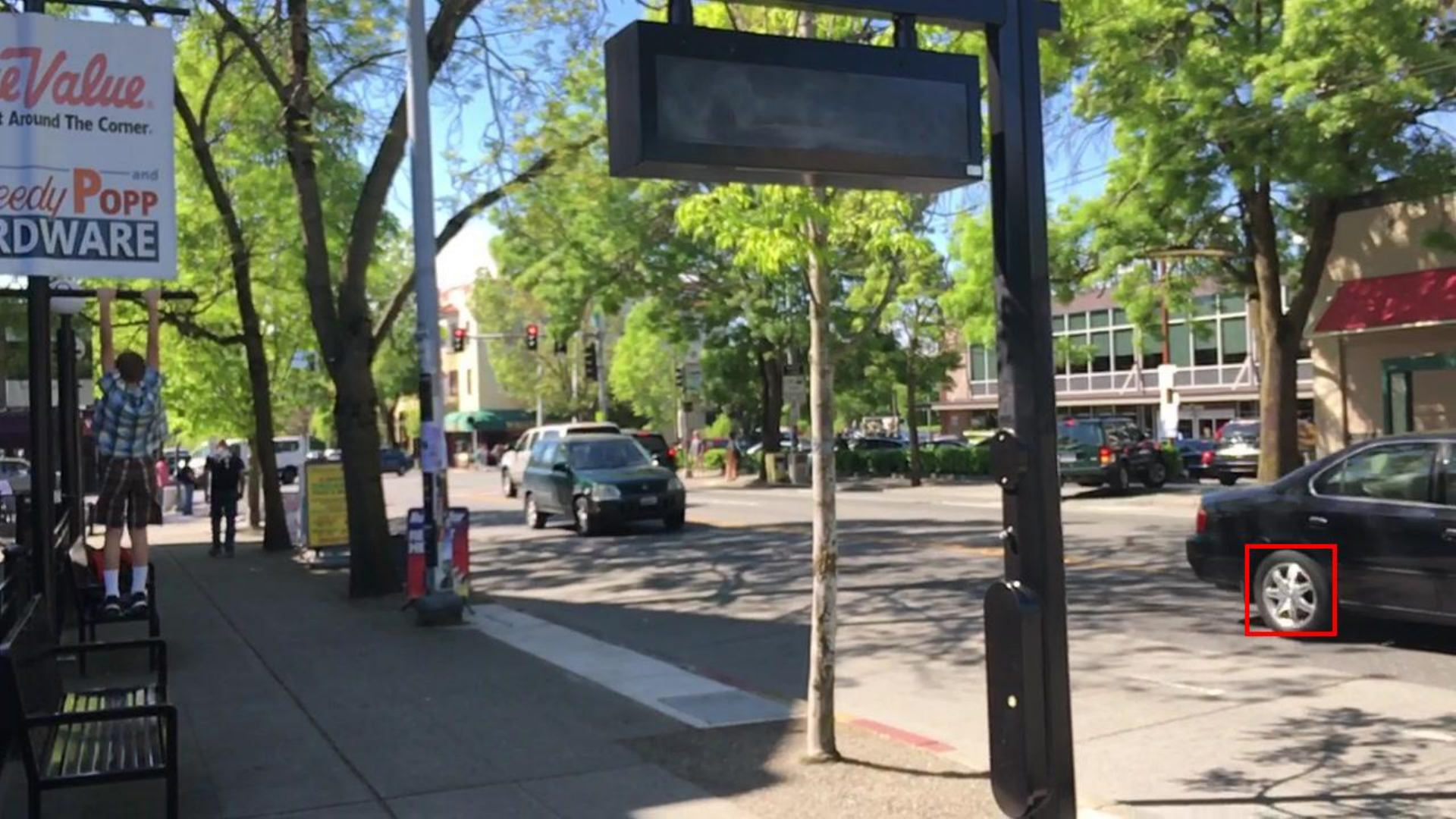}
\\

\end{tabular}
\end{adjustbox}
\hspace{-0.15cm}
\begin{adjustbox}{valign=t}
\begin{tabular}{cccccc}
\includegraphics[width=0.1\textwidth]{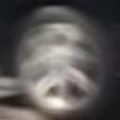} \hspace{-1mm}  &
\includegraphics[width=0.1\textwidth]{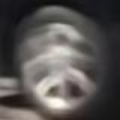} \hspace{-1mm}  &
\includegraphics[width=0.1\textwidth]{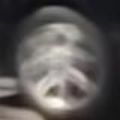} \hspace{-1mm} &
\includegraphics[width=0.1\textwidth]{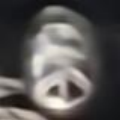} \hspace{-1mm} &
\includegraphics[width=0.1\textwidth]{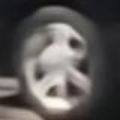} \hspace{-1mm}
\\
Blur \hspace{-1mm} &
EDVR \hspace{-1mm} &
MPRNet \hspace{-1mm} &
TSP \hspace{-1mm} &
SFE \hspace{-1mm}
\\
\includegraphics[width=0.1\textwidth]{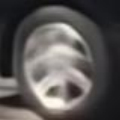} \hspace{-1mm} &
\includegraphics[width=0.1\textwidth]{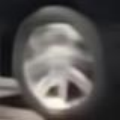} \hspace{-1mm} &
\includegraphics[width=0.1\textwidth]{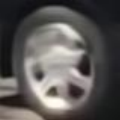} \hspace{-1mm} &
\includegraphics[width=0.1\textwidth]{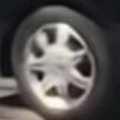} \hspace{-1mm} &
\includegraphics[width=0.1\textwidth]{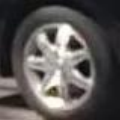}  \hspace{-1mm} 
\\ 
BasicVSR \hspace{-1mm} &
FGST \hspace{-1mm} &
DSTNet \hspace{-1mm} &
Ours  \hspace{-1mm} &
GT \hspace{-1mm}
\\
\end{tabular}
\end{adjustbox}
\end{tabular}
\caption{Visual comparison on DVD \cite{DVD} dataset. The deblurring effect of our proposed model is significantly better.}
\label{fig_dvd}
\end{figure*}

\begin{figure*}[t]
\scriptsize
\centering
\begin{tabular}{ccc}
\hspace{-0.36cm}
\begin{adjustbox}{valign=t}
\begin{tabular}{cccccc}
\includegraphics[width=0.25\textwidth]{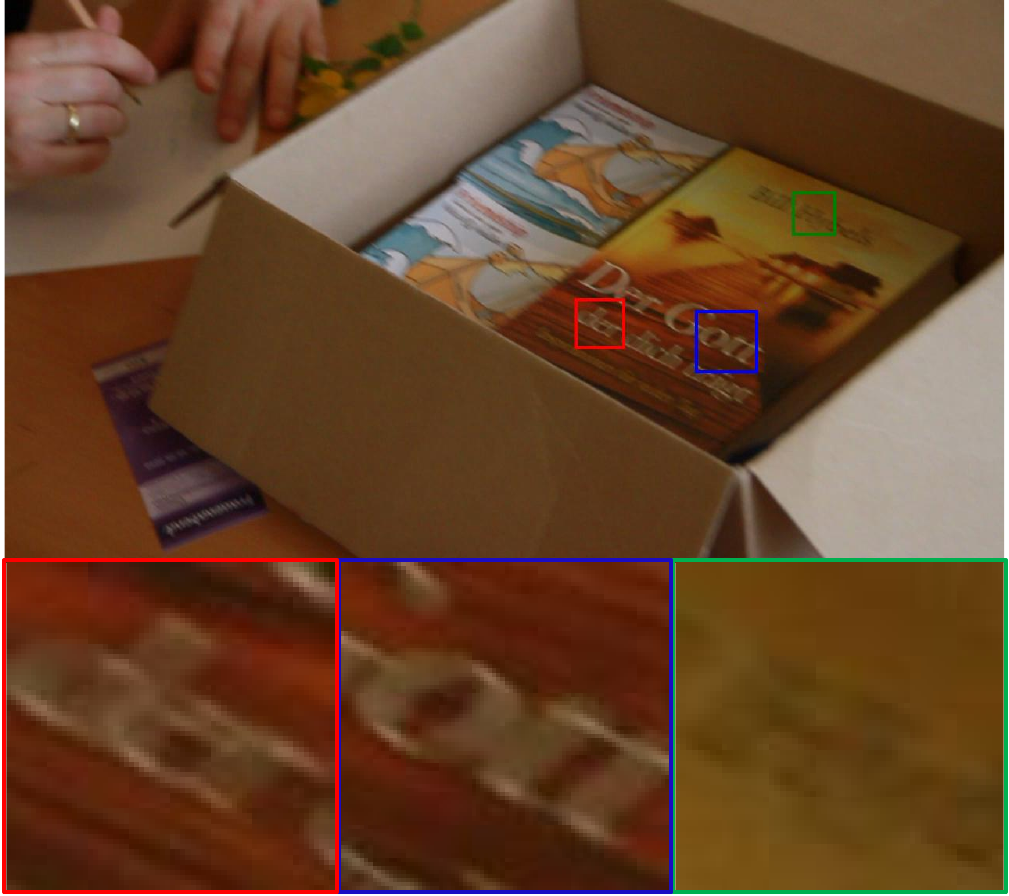} \hspace{-3mm}  &
\includegraphics[width=0.25\textwidth]{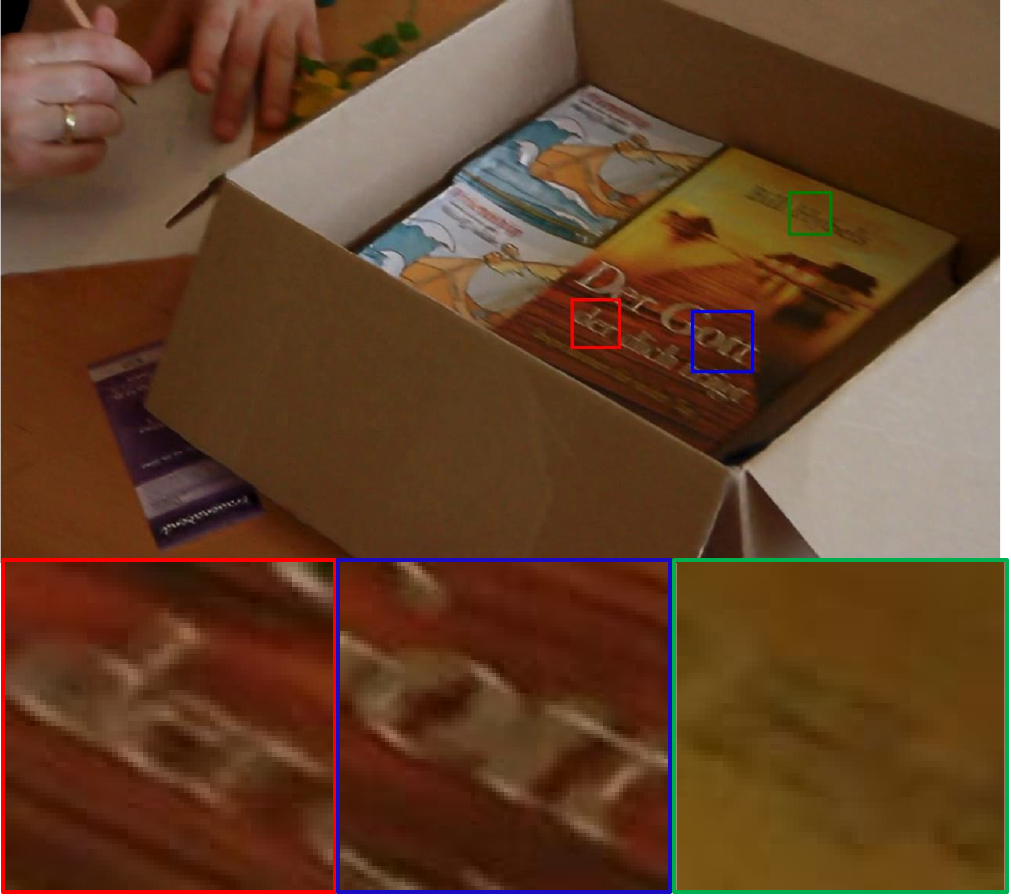} \hspace{-3mm}  &
\includegraphics[width=0.25\textwidth]{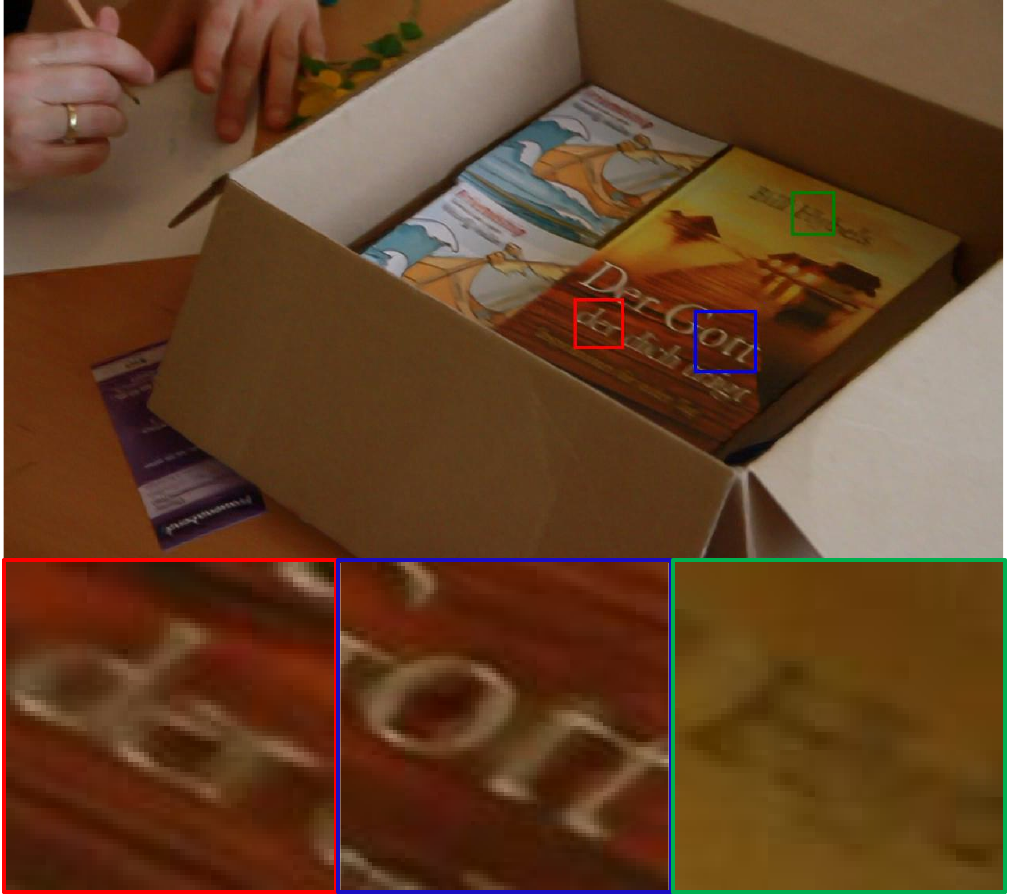} \hspace{-3mm} &
\includegraphics[width=0.25\textwidth]{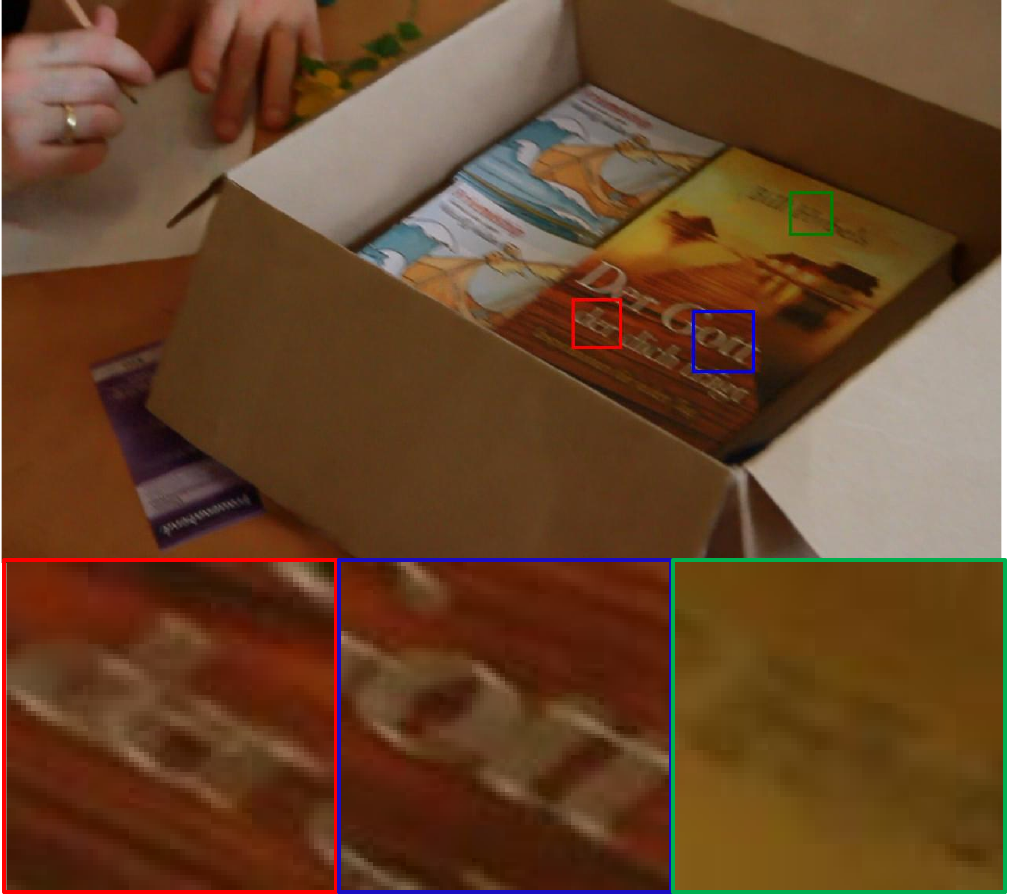} \hspace{-3mm} &
\\
Blur~ \hspace{-3mm} &
DVD~\cite{DVD} \hspace{-3mm} &
EDVR~\cite{2019EDVR} \hspace{-3mm} &
MPRNet~\cite{2021MPRNet} \hspace{-3mm} &
\\
\includegraphics[width=0.25\textwidth]{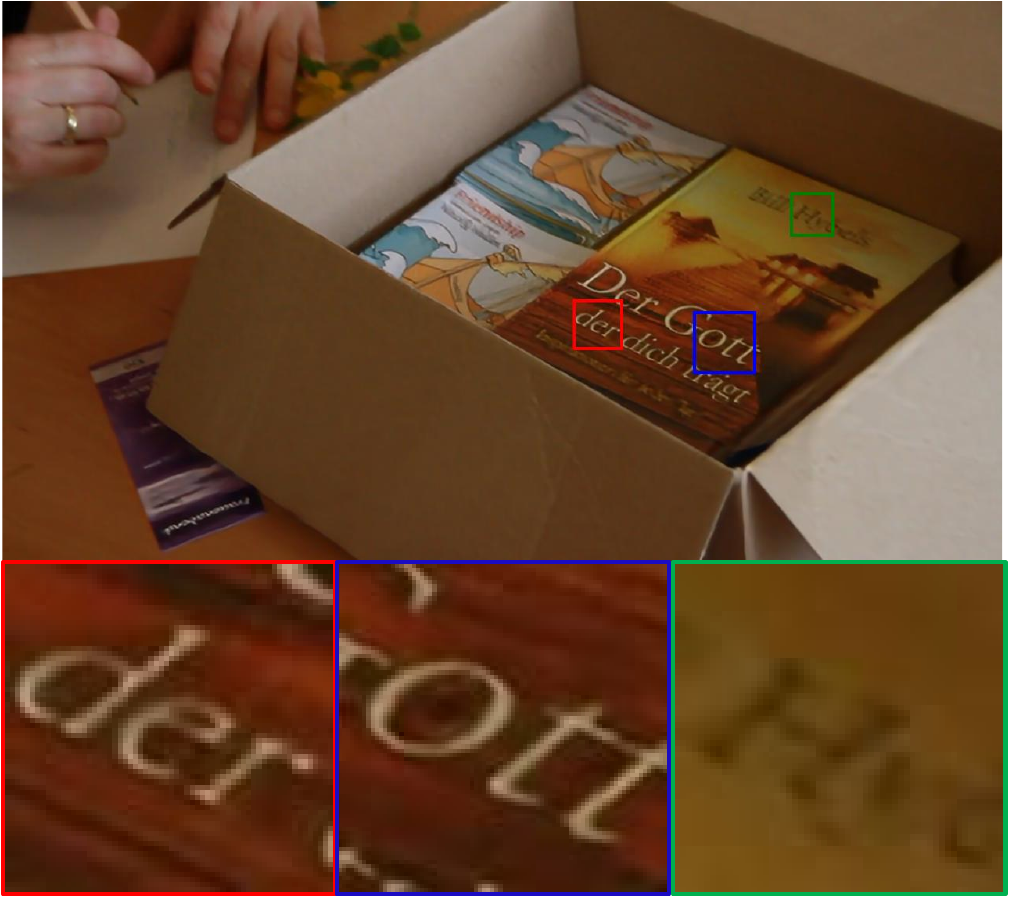} \hspace{-3mm} &
\includegraphics[width=0.25\textwidth]{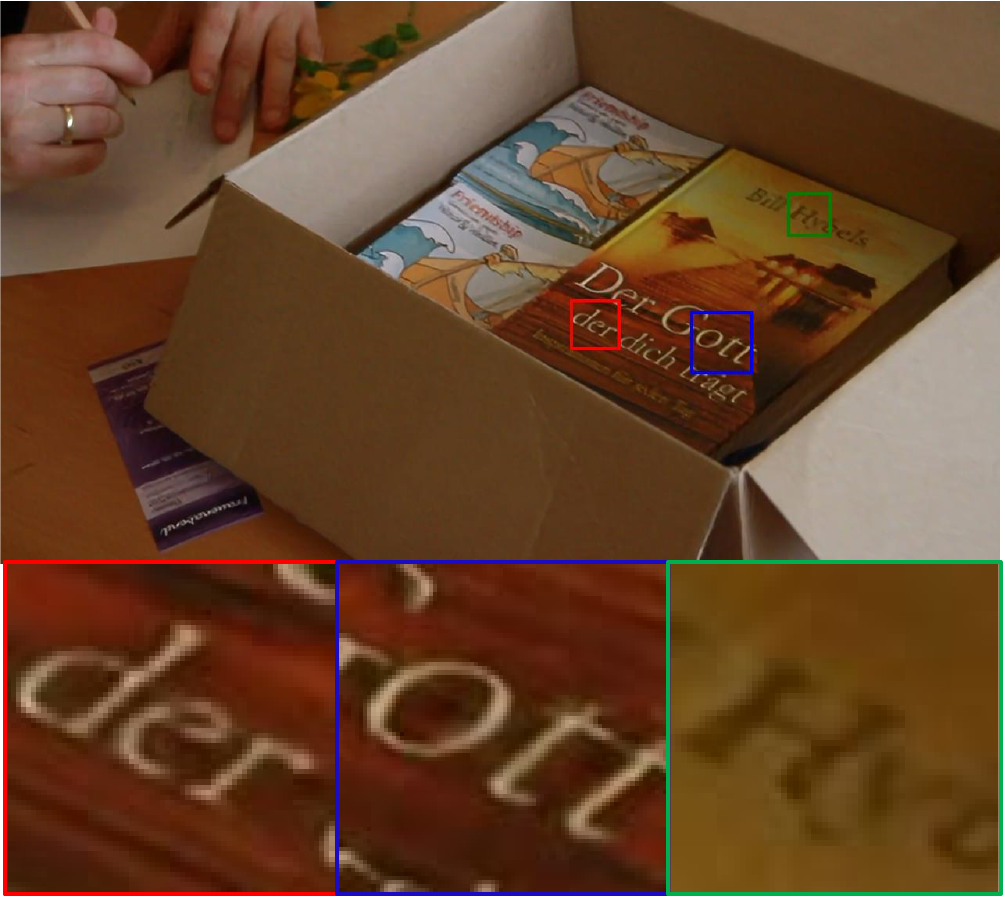} \hspace{-3mm} &
\includegraphics[width=0.25\textwidth]{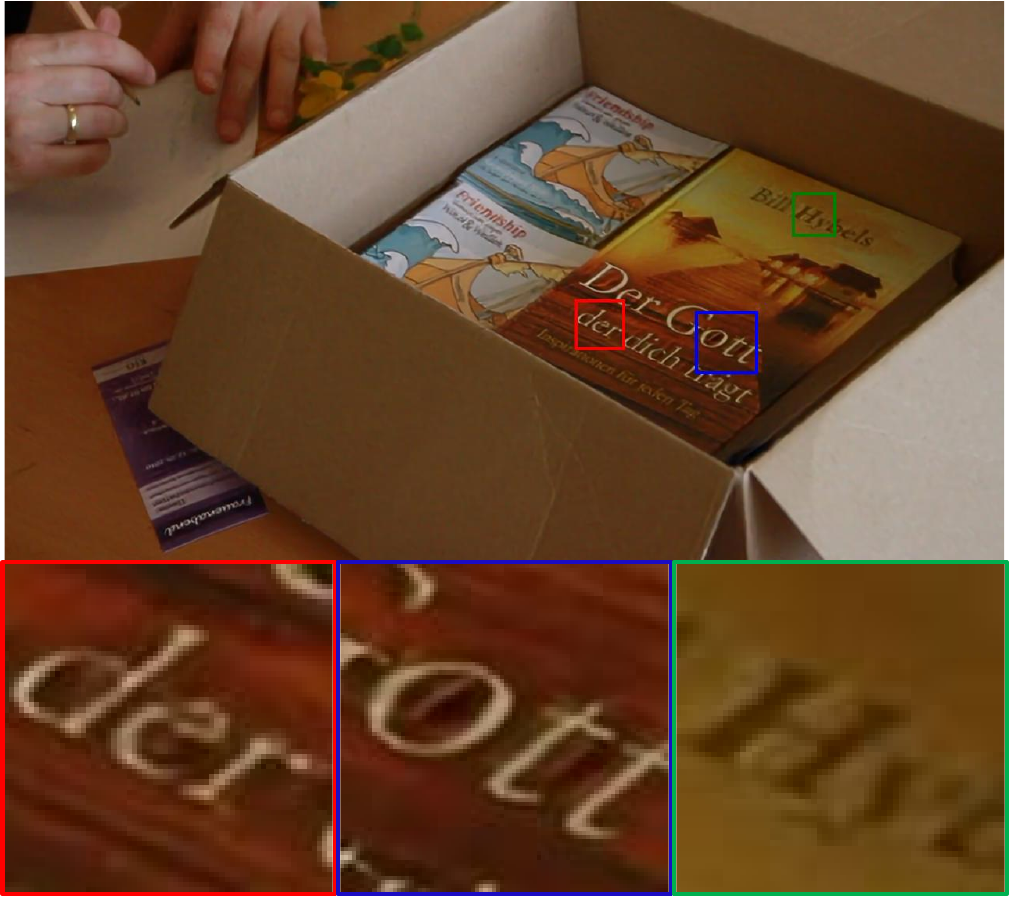} \hspace{-3mm} &
\includegraphics[width=0.25\textwidth]{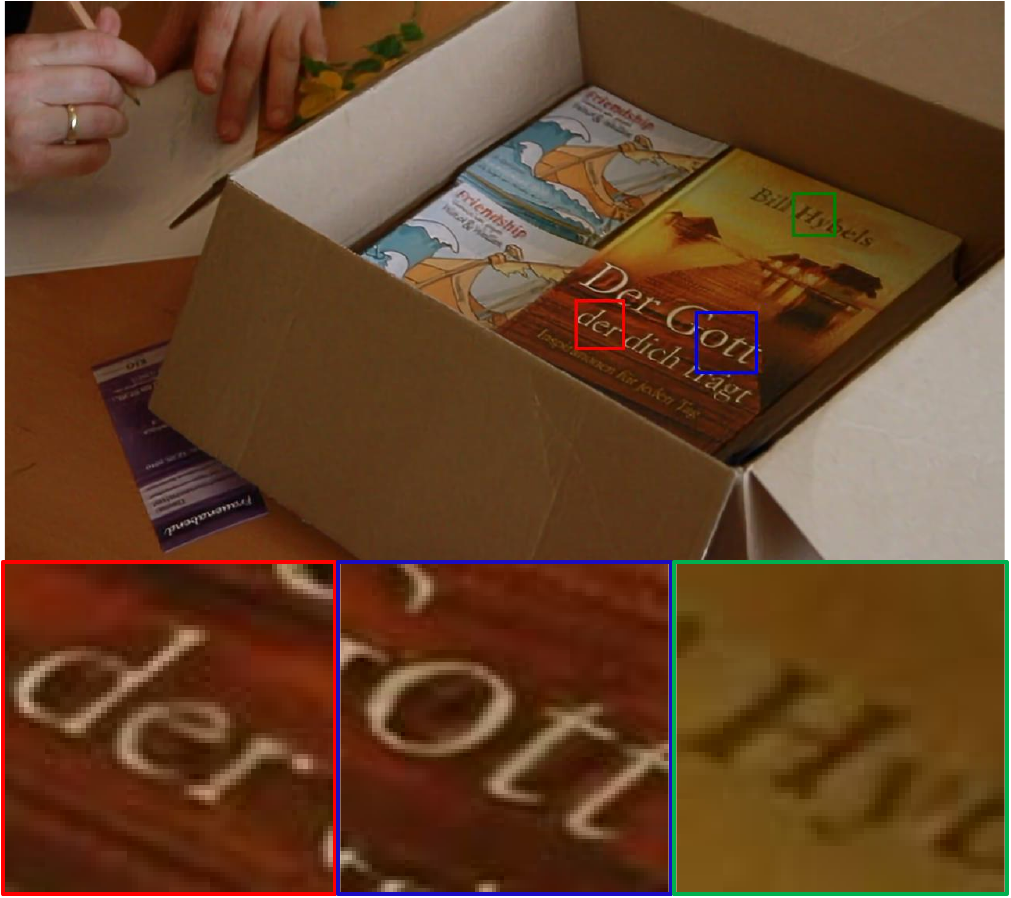} \hspace{-3mm} &
\\ 
TSP~\cite{2020CDVDTSP} \hspace{-3mm} &
SFE~\cite{2020DVDSFE} \hspace{-3mm} &
DSTNet~\cite{DSTNet} \hspace{-3mm} &
VD-Diff (Ours)~  \hspace{-3mm} &
\\
\end{tabular}
\end{adjustbox}
\end{tabular}
\caption{Visual results of VD-Diff and SOTA methods on the real blurry videos of \cite{DVD}.}
\label{fig_real}
\end{figure*}

\begin{figure}[t!]
  \centering
   \includegraphics[width=\linewidth]{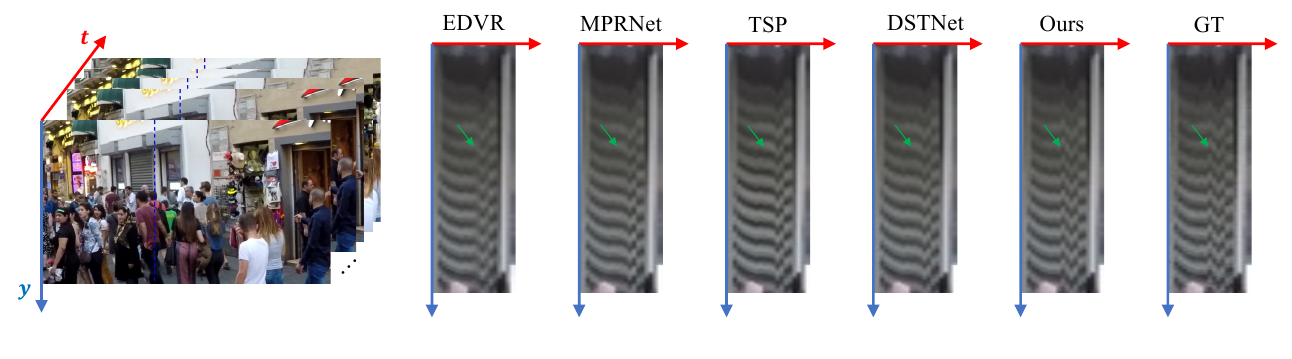}
   \caption{Comparison of various methods in terms of time dimension.}
   \label{t_dimension}
\end{figure}

\begin{figure}[t!]
  \centering
   \includegraphics[width=\linewidth]{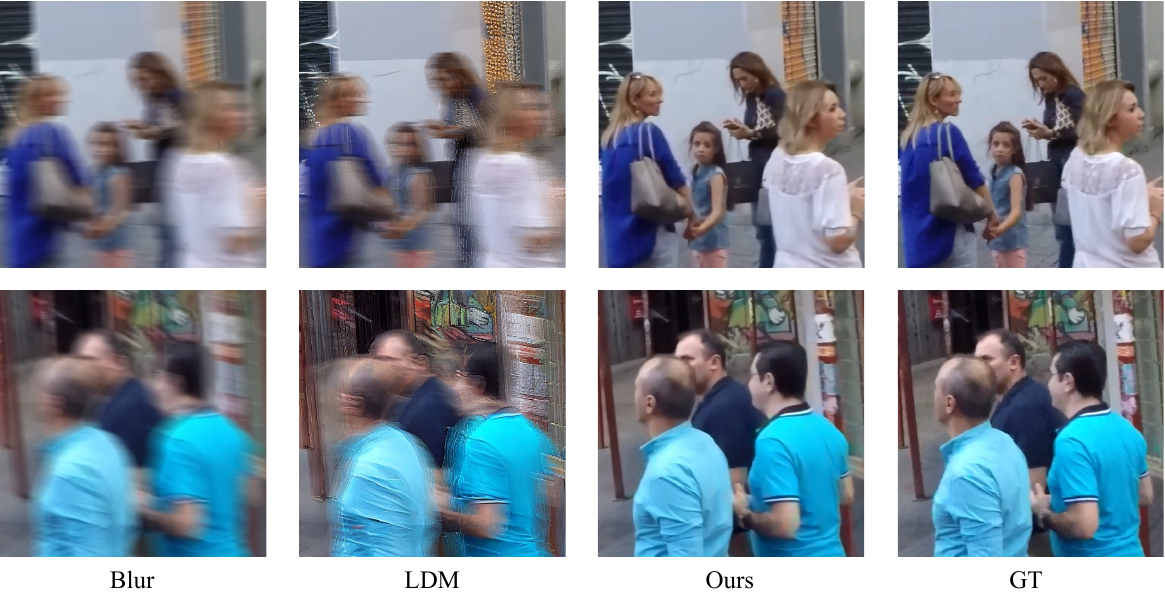}
   \caption{Qualitative comparison with pre-trained large diffusion models \cite{rombach2022high}.}
   \label{comp_sd}
\end{figure}

\section{Experiments}
\subsection{Datasets}
\noindent \textbf{GoPro.} The GoPro \cite{GoPro} benchmark is composed of over 3,300 blurry-sharp image pairs of dynamic scenes. It is obtained by a high-speed camera. The training and testing subsets are split in proportional to 2:1.

\noindent \textbf{DVD.} The DVD \cite{DVD} dataset consists of 71 videos with 6,708 blurry-sharp image pairs. It is divided into train/test subsets with 61 videos (5,708 image pairs) and 10 videos (1,000 image pairs). DVD is captured with mobile phones and DSLR at a frame rate of 240 FPS.

\noindent \textbf{BSD.} The Beam-Splitter Deblurring Dataset (BSD) \cite{2020ESTRNN} consists of 240 videos with 24k blurry-sharp image pairs. Cross-validation experiments between real-world and synthetic datasets demonstrate the high generality of the BSD dataset.

\noindent \textbf{Real-World Videos.} We evaluate the generality of our proposed approach on the real blurry datasets \cite{2012realworldvideo}. Due to the inaccessibility of the ground truth, we only compare VD-Diff with other models on visual results.

\subsection{Implementation Details}
We implement VD-Diff in PyTorch \cite{paszke2019pytorch} using a single NVIDIA RTX A6000 GPU. The training process takes 7 days and 10 hours. 
We set the batch size as 2, the learning rate as $8\times 10^{-5}$ and weight decay as $4\times 10^{-5}$. Patches at the size of 256×256 cropped from training frames are fed into the models. The sequence length is set to 15 in training and 16 in testing. Since we use diffusion within a highly compressed latent space, the denoising iteration steps $T$ only needs 4. We set $C$ to 64 and $C'$ to 256, respectively. In the WADT module, we set $N_1$ and $N_2$ to 7 and 2, respectively, to balance the calculation amount and the deblurring effect. The horizontal and vertical flips are performed for data augmentation.

\subsection{Quantitative Results}
\begin{figure}[t!]
  \centering
   \includegraphics[width=\linewidth]{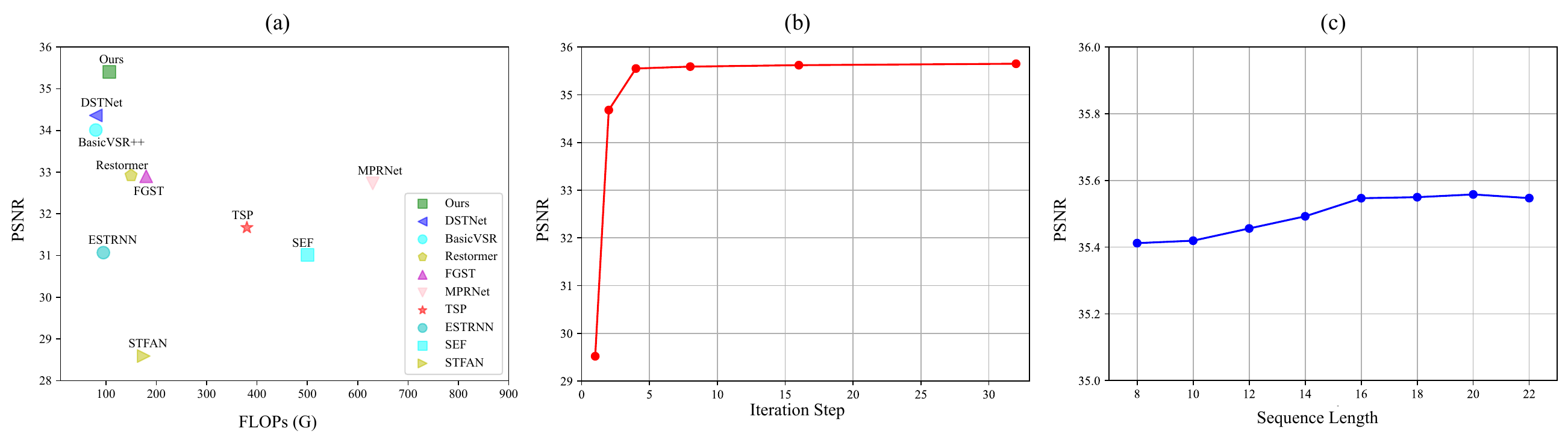}
   \caption{(a) Floating point operations (FLOPs) \emph{vs} video deblurring performance on the GoPro \cite{GoPro} dataset. Our model achieves favorable results in terms of accuracy and FLOPs. (b) Ablation study of the number of iteration steps. (c) Ablation study of the sequence length during inference.}
   \label{threeFig}
\end{figure}

We compare our proposed VD-Diff with other SOTA methods, as shown in Tables \ref{GoProQuantitative}, \ref{DVDQuantitative}, and \ref{BSDQuantitative}. 
As can be seen, our method outperforms SOTA methods on GoPro \cite{GoPro} and DVD \cite{DVD} benchmark datasets. 
Specifically, as shown in Table \ref{GoProQuantitative}, Our method outperforms DSTNet-L, RVRT, and FGST by 0.50dB, 0.63dB, and 2.65dB in PSNR on the GoPro \cite{GoPro} dataset, respectively. 
Our model achieves SOTA performance on the GoPro \cite{GoPro} dataset. Table \ref{DVDQuantitative} shows that our proposed method generates the deblurred videos with higher PSNR and SSIM values on the DVD \cite{DVD} dataset. 
%
As shown in Table \ref{BSDQuantitative}, we also achieve SOTA on the BSD \cite{2020ESTRNN} dataset.
As seen from Table \ref{parameters}, our model has a relatively low parameter size and less inference time than other models. Although our model's parameter size and inference time are slightly higher than DSTNet and BasicVSR++, the PSNR metric on the GoPro \cite{GoPro} dataset is at least 1.3dB higher than the two models. 
As shown in Fig. \ref{threeFig}(a), our model achieves favorable results in terms of accuracy and FLOPs. 
The above results show the effectiveness of our model.

\subsection{Qualitative Results}
We provide visual comparisons on GoPro \cite{GoPro}, DVD \cite{DVD}, and Real-World Video datasets \cite{DVD} as shown in Fig. \ref{fig_gopro}, \ref{fig_dvd}, and \ref{fig_real}, respectively. Some models sacrifice texture and details to maintain overall structure, resulting in over-smooth videos that lose high-frequency detail. Other models ignore global information to preserve high-frequency detail, introducing redundant speckle textures and artifacts that do not match the real video in fast-motion video. In contrast, our VD-Diff has the best visual effect as it can leverage WADT to model long-range dependencies to capture low-frequency information and employ DM to supplement valuable high-frequency information. Therefore, as seen from the visual comparison, our model can restore texture and details while maintaining the overall framework of the video.

Besides, we conduct a qualitative comparison in terms of the time dimension, as shown in Fig. \ref{t_dimension}. As can be seen, previous models such as EDVR \cite{2019EDVR}, MPRNet \cite{2021MPRNet}, TSP \cite{2020CDVDTSP}, and DSTNet \cite{DSTNet} all exhibit unclear outlines, indicating that they cannot guarantee temporal consistency. In contrast, our method effectively maintains temporal consistency.

Furthermore, we conduct a comparison with the direct application of a pre-trained large diffusion model (\emph{i.e.}, LDM \cite{rombach2022high}) to video deblurring, as shown in Fig. \ref{comp_sd}. We notice that the videos deblurred by LDM contain a significant number of artifacts and inconsistencies with the content of the original videos, which is attributed to motion blur present in the blurred videos. In contrast, our method leverages the diffusion model and WADT to maximally restore the details in the video while preserving the video content.

\begin{table*}[t]
\begin{center}
\small
\setlength{\tabcolsep}{2.5pt}
\scalebox{0.9}{
\hspace{-5mm}
\begin{tabular}{lccccccccccc}
\hline
\hline
\rowcolor[gray]{0.9} Method &DVD \cite{DVD} &EDVR \cite{2019EDVR}  &STFAN \cite{2019STFAN} &ESTRNN \cite{2020ESTRNN}   &MPRNet \cite{2021MPRNet} &TSP \cite{2020CDVDTSP} \\
\hline
\hline
PSNR~$\textcolor{black}{\uparrow}$ &27.31 &26.83 &28.59 &31.07 &32.73 &31.67\\
SSIM~$\textcolor{black}{\uparrow}$ &0.8255 &0.8426 &0.8608 &0.9023 &0.9366 &0.9279\\
\hline
\hline
\rowcolor[gray]{0.9} BasicVSR \cite{basicvsr++Generalization} &NAFNet \cite{NAFNet} &FGST \cite{FGST}  &DSTNet \cite{DSTNet}    &DSTNet-L \cite{DSTNet}   &RVRT \cite{2022rvrt}   & \textbf{Ours}   \\
\hline
\hline 
34.01 &33.71 &32.90 &34.16 &35.05 &34.92 &\textbf{35.55}\\ 
0.9520 &0.9668  &0.9610 &0.9679 &0.9733 &0.9738 &\textbf{0.9752}\\
\hline
\hline
\end{tabular}
}
\caption{VD-Diff achieves SOTA results on the GoPro dataset \cite{GoPro}.}
\label{GoProQuantitative}
\end{center}\vspace{-5.5mm}
\end{table*}

\begin{table*}[t]
\begin{center}
\setlength{\tabcolsep}{2.5pt}
\scalebox{0.9}{
\begin{tabular}{lccccccccccc}
\hline
\hline
\rowcolor[gray]{0.9} Method &DVD \cite{DVD}  &EDVR \cite{2019EDVR}  &STFAN \cite{2019STFAN} &ESTRNN \cite{2020ESTRNN} &TSP \cite{2020CDVDTSP}    &SEF \cite{2020DVDSFE} \\
\hline
\hline
PSNR~$\textcolor{black}{\uparrow}$ &30.01 &28.51 &31.15 &32.01 &32.13 &31.71\\
SSIM~$\textcolor{black}{\uparrow}$ &0.8877 &0.8637 &0.9049 &0.9162 &0.9268 &0.9160\\
\hline
\hline
\rowcolor[gray]{0.9} GSTA \cite{2021GSTA}  &MPRNet \cite{2021MPRNet} &ARVo \cite{2021ARVo}   &FGST \cite{FGST}   &DSTNet \cite{DSTNet}   &RVRT \cite{2022rvrt}      & \textbf{Ours}   \\
\hline
\hline
32.53 &32.73 &32.80  &33.36 &33.79 &34.30 &\textbf{34.56}\\
0.9468 &0.9366 &0.9352 &0.9500 &0.9615 &0.9655 &\textbf{0.9684}\\
\hline
\hline
\end{tabular}
}
\caption{VD-Diff achieves SOTA results on the DVD dataset \cite{DVD}.}
\label{DVDQuantitative}
\end{center}\vspace{-5.5mm}
\end{table*}

\begin{table*}[t]
\begin{center}
\setlength{\tabcolsep}{2.5pt}
\scalebox{0.9}{
\begin{tabular}{lccccccccccc}
\hline
\hline
\rowcolor[gray]{0.9} Method &DVD \cite{DVD}  &DTBN \cite{hyun2017online}  &Nah et al. \cite{nah2019recurrent} &ESTRNN \cite{2020ESTRNN} &TSP \cite{2020CDVDTSP}    &DSTNet \cite{DSTNet} & \textbf{Ours}\\
\hline
\hline
PSNR~$\textcolor{black}{\uparrow}$ &29.95 &31.84 &33.00 &33.36 &32.84 &34.45 &\textbf{35.61}\\
SSIM~$\textcolor{black}{\uparrow}$ &0.8692 &0.9170 &0.9330 &0.9370 &0.9398 &0.9548 &\textbf{0.9758}\\
\hline
\hline
\end{tabular}
}
\caption{VD-Diff achieves SOTA results on the BSD dataset \cite{2020ESTRNN}.}
\label{BSDQuantitative}
\end{center}\vspace{-5.5mm}
\end{table*}

\begin{table*}[t]
\begin{center}
\small
\setlength{\tabcolsep}{2.5pt}
\scalebox{0.9}{
\begin{tabular}{l c c c c c c c c c c c}
\hline
\hline
\rowcolor[gray]{0.9} Method &DVD &EDVR  &TSP    &FGST   &DSTNet &BasicVSR++ &RVRT  &DSTNet-L    & \textbf{Ours}\\
\hline
\hline
Parameters (M)~ &15.30 &23.60 &16.19 &9.70 &7.45 &9.46  &13.60 &16.96 &11.93\\
Running time (s)~ &0.88 &2.70 &1.90 &0.72 &0.08 &0.10 &0.41 &0.28 &0.31\\
\hline
\hline
\end{tabular}
}
\vspace{0mm}
\caption{Quantitative evaluations of the video deblurring methods regarding network parameters and running time. The running time is obtained on a machine with a single NVIDIA RTX A6000 GPU. The size of the test images is 1280 × 720 pixels.}
\label{parameters}
\end{center}
\end{table*}

\begin{table*}[h!]
\small
\centering
\scalebox{0.9}{
\begin{tabular}{l|c|c|c|c|c|c}   
\hline
\hline
\multirow{2}*{Variants} & \multicolumn{4}{c|}{\cellcolor[gray]{0.9} Components} &\multicolumn{2}{c}{\cellcolor[gray]{0.9} Metrics} \\  
\cline{2-7}
 & \cellcolor[gray]{0.9} Diffusion Model & \cellcolor[gray]{0.9} Wavelet & \cellcolor[gray]{0.9} Joint Training & \cellcolor[gray]{0.9} WBPF & \cellcolor[gray]{0.9} PSNR $\uparrow$ & \cellcolor[gray]{0.9} SSIM $\uparrow$   \\
\hline
\hline
\emph{w/o} DM &  \textcolor{eccvblue}{\usym{2717}} & \usym{2713} & \usym{2713} & \usym{2713}    &34.38  & 0.9684  
\\
\emph{w/o} Wavelet & \usym{2713} &   \textcolor{eccvblue}{\usym{2717}} & \usym{2713} & \usym{2713}   & 34.75  &  0.9709
\\
\emph{w/o} Joint Training& \usym{2713} & \usym{2713}  &   \textcolor{eccvblue}{\usym{2717}} & \usym{2713}  &34.92   &  0.9717
\\
\emph{w/o} WBPF & \usym{2713} & \usym{2713}  & \usym{2713}  &   \textcolor{eccvblue}{\usym{2717}}   &33.58   &  0.9344
\\
Full model & \usym{2713} &  \usym{2713}  &  \usym{2713}  &  \usym{2713}  & \textbf{35.55}  &  \textbf{0.9752}
  \\
\hline
\hline
\end{tabular}
} 
\caption{Ablation experiments on various variants of VD-Diif on the GoPro \cite{GoPro} dataset, and the best quantitative results are marked in \textbf{bold}.}
\label{tab_ab}
\end{table*}

\subsection{Ablation Study}
This section explores the significance of each key component of our proposed VD-Diff. All variants are tested on the GoPro \cite{GoPro} dataset.

{\bf Effect of Diffusion Model.} 
To validate the effectiveness of the diffusion model, we design a variant that does not utilize DM to generate prior features but solely relies on WADT for video deblurring, named as \emph{w/o} DM. As seen in Table \ref{tab_ab}, without employing DM, the deblurring performance of the network significantly declines, with a decrease of 1.17 in PSNR, which indicates that DM can provide valuable prior features to enhance the deblurring performance.

{\bf Effect of Wavelet Transform.} We conduct ablation experiments to study the impact of wavelet transform on VD-Diff. We replace wavelet transform and inverse wavelet transform with bilinear downsample and upsample to evaluate the effect of wavelet transform, named \emph{w/o} Wavelet. We adopt the same training strategy as VD-Diff. Table \ref{tab_ab} shows \emph{w/o} Wavelet decreases by 0.8dB regarding the PSNR metrics. This experiment shows 
the wavelet transform method for feature extraction is better than the general upsampling and downsampling methods.

{\bf Effect of Joint Training Strategy.} In Sec. \ref{sec:TrainingStrategy}, we analyze the importance of joint training. To verify our analysis, we remove joint training stage \emph{i.e.}, Training Stage Three, as shown in Table \ref{tab_ab}.
We notice that VD-Diff is significantly better than \emph{w/o} Joint Training with an improvement of 0.63dB in terms of PSNR metrics, which demonstrates that joint training can improve the collaborative working ability of DM and WADT, thereby improving the deblurring effect.

{\bf Effect of Wavelet-based Bidirectional Propagation Fuse.} 
To investigate the effect of Wavelet-based Bidirectional Propagation Fuse (WBPF), we design a variant model by removing the WBPF in WADT, termed as \emph{w/o} WBPF. 
Table \ref{tab_ab} shows that WBPF can effectively transmit useful information between frames, improving the model's ability to capture spatio-temporal information. 

{\bf Effect of Iteration Steps.} To explore the impact of the number of iterations on the diffusion model, we set different iteration steps to train the model and then compare the PSNR metrics on the GoPro dataset. The results are shown in Fig. \ref{threeFig}(b). The value of PSNR increases as the number of iteration steps increases, but when the number of iterations is greater than 4, the value of PSNR almost remains stable. It shows that in a highly compressed latent space, just setting the number of iteration steps to 4 can generate suitable prior features.

{\bf Effect of Sequence Length.} Intuitively, the sequence length will affect the deblurring effect of our proposed model. If the sequence length is too short, the model fails to capture essential information from distant frames. Conversely, if the sequence length is too long, the computational load becomes excessive, potentially causing error accumulation. To investigate this, we experimented with various sequence lengths. As shown in Fig. \ref{threeFig}(c), the PSNR value increases steadily with the number of frames. However, after 16 frames, the calculation becomes very large as the sequence length increases, and the PSNR does not change significantly. Therefore, we set the sequence length to 16 to reach a near-optimal point between the deblurring effect and computational reduction.

\section{Conclusion}
In this paper, we propose a brand-new method that innovatively integrates the diffusion model with the Transformer for video deblurring.
Specifically, the diffusion model generates prior features containing high-frequency information that conforms to the ground truth (GT) distribution. Meanwhile, the Wavelet-Aware Dynamic Transformer captures low-frequency information and utilizes the prior features generated by the diffusion model to supplement high-frequency information, thereby enhancing the quality of the deblurred video.
Comprehensive experimental results show that our proposed method outperforms the state-of-the-art methods.

\section*{Acknowledgements}
This work was supported in part by National Social Science Foundation Major Project "Research on Virtual Restoration of Tang and Song Painting Colors and Construction of Traditional Color Resource Library from a Digital Perspective" (19ZDA046), Zhejiang Province Program (2022C01222, 2023C03199, 2023C03201), the National Program of China (62172365, 2021YFF0900604,\\ 19ZDA197), Ningbo Science and Technology Plan Project (2022Z167, 2023Z137), and MOE Frontier Science Center for Brain Science \& Brain-Machine Integration (Zhejiang University).

%
%
\bibliographystyle{splncs04}
\bibliography{main}
\end{document}